\documentclass{Interspeech}

\interspeechcameraready

\title{Who Gets the Mic? Investigating Gender Bias in the Speaker Assignment of a Speech-LLM}

\author[affiliation={1}]{Dariia}{Puhach}
\author[affiliation={3}]{Amir H.}{Payberah}
\author[affiliation={2}]{Éva}{Székely}

\affiliation{Department of Linguistics and Philology}{Uppsala University}{Sweden}
\affiliation{Division of Software and Computer Systems}{KTH Royal Institute of Technology}{Sweden}
\affiliation{Department of Speech, Music, and Hearing}{KTH Royal Institute of Technology}{Sweden}
\email{daschablume@gmail.com, payberah@kth.se, szekely@kth.se}

\usepackage{comment}

\begin{document}

\maketitle

\begin{abstract} 
Similar to text-based Large Language Models (LLMs), Speech-LLMs exhibit emergent abilities and context awareness. However, whether these similarities extend to gender bias remains an open question. This study proposes a methodology leveraging speaker assignment as an analytic tool for bias investigation. Unlike text-based models, which encode gendered associations implicitly, Speech-LLMs must produce a gendered voice, making speaker selection an explicit bias cue.
We evaluate Bark, a Text-to-Speech (TTS) model, analyzing its default speaker assignments for textual prompts. If Bark’s speaker selection systematically aligns with gendered associations, it may reveal patterns in its training data or model design. To test this, we construct two datasets: (i) Professions, containing gender-stereotyped occupations, and (ii) Gender-Colored Words, featuring gendered connotations. While Bark does not exhibit systematic bias, it demonstrates gender awareness and has some gender inclinations.

\end{abstract}

\section{Introduction}

Bias in generative AI models is an active area of research, with studies showing that Large Language Models (LLMs) reflect and sometimes amplify societal biases, including gender bias  \cite{zhao2024genderbiaslargelanguage, Kotek_2023, knowledge-distillation}. 
Speech-LLMs differ in that they do not just imply gender, they must produce a signal inherently carrying gendered associations, even when ambiguous. This requirement makes speaker selection in Speech-LLMs a uniquely explicit lens for studying bias, yet existing research has not fully explored its potential as a diagnostic tool.

Speech-LLMs have advanced significantly in their ability to model language from raw audio \cite{lakhotia-etal-2021-generative} and -- adapted to Text-to-Speech (TTS) tasks -- generate expressive, rich prosody in speech without relying on textual corpora or Automatic Speech Recognition (ASR) ~\cite{betker2023better, lajszczak2024base}.
One example of a Speech-LLM-based TTS system is Bark by Suno\footnote{https://github.com/suno-ai/bark}. It is a quite popular model, used as a baseline for end-to-end English speech conversion~\cite{tathe2024end} and synthetic data generation for low-resource ASR~\cite{kamble2024customdataaugmentationlow}. Efforts have also been made to enhance its output accuracy~\cite{schumacher2023enhancing}. Compared to the traditional TTS system VITS \cite{kim2021conditional}, Bark exhibits lower intelligibility and robustness but higher spontaneity~\cite{wang2024evaluatingtexttospeechsynthesislarge}, along with significant acoustic diversity in voice, gender, and emotion~\cite{madspeech}.

Since speaker selection is an inherent feature of speech synthesis, it can provide a direct and measurable way to examine gender associations in model behavior. Unlike text-based evaluations, which rely on indirect inference (e.g., whether a model assigns gendered pronouns to professions), speaker assignment requires an explicit gender representation for each output, making it a potentially powerful indicator of underlying biases in training data.

This study introduces a methodology using speaker selection as an analytic tool to examine bias in Speech-LLMs. Since Bark automatically assigns a speaker when no speaker prompt is provided, we analyze its default gender assignments across different textual inputs. We hypothesize that systematic preferences—assigning speaker voices based on gendered associations reflecting societal stereotypes—may stem from biases in training data or model design choices. While a single speaker assignment may not constitute harmful bias in and of itself, the systematic distribution of gendered outputs can reveal trends that may otherwise be difficult to surface in generative models.

To explore these questions, we construct two datasets: (i) the {\em Professions} dataset, containing sentences with stereotypically male and female professions, and (ii) the {\em Gender-Colored Words (GCW)} dataset, consisting of sentences with gender-associated (colored) words\footnote{The code and datasets this study are publicly available at https://github.com/daschablume/speech-gender-bias.}. We find that Bark accurately associates gender-conforming names with gender but shows more diversity in speaker assignments for both Professions and GCW. However, a small subset of words in both datasets still lead to gender preference, not always aligning with the bias. Additionally, our analysis confirms that Bark infers gender information at one of its layers.

\section{Method}
 We create two types of datasets to test gender bias in Bark: (i) the {\em Professions} dataset and (ii) the {\em Gender-Colored Words (GCW)} dataset. Each sentence from both datasets is input into Bark 10 times to group the results and mitigate randomness. The audio outputs are then classified using the Speaker Gender Recognition (SGR) model trained by~\cite{Székely2023ProsodycontrollableGS}, which employs pre-trained wav2vec 2.0~\cite{baevski2020wav2vec} speech representations. The SGR model is trained on the LibriTTS dataset~\cite{libritts}, with a subset of male speakers selected to ensure a balanced gender distribution. The output of SGR is binary, providing male and female probabilities for an utterance. However, since Bark occasionally generated low-quality or incomplete outputs, manual listening tests were conducted to ensure the accuracy of gender classification.

In the rest of this section, we first describe the two datasets, then define the baseline, and finally explain the additional testing process with a speaker prompt.

 \subsection{Professions Dataset}  
For the Professions dataset, we drew inspiration from the WinoBias dataset \cite{winobias}, which defines stereotypical professions based on the US occupational statistics. WinoBias includes sentences containing two professions, one stereotypically male and one stereotypically female--alongside a pronoun. The goal of a text-based model is to resolve the pronoun. 
To adapt this for Bark, we rewrote the sentences in the first person, such as: ``\textit{I work as a developer. I argued with the designer}''. This approach resulted in 26 sentences, each featuring a unique profession. 

\subsection{GCW Dataset}
``Gender-colored words" (GCW) refer to words with gender connotations. This includes both strongly gendered terms like ``bloke" and more subtly associated words like ``tutu", which refers to a skirt worn by female dancers\footnote{https://dictionary.cambridge.org/dictionary/english/tutu}.
For creation of GCW dataset, we used the data from the Word Association Graph\footnote{https://github.com/Yupei-Du/bias-in-wat}~\cite{du-etal-2019-exploring}. This graph contains gender-associated words, with ``gentleman'' and ``lady'' at opposite extremes. However, the distribution is skewed, with only 228 ``male'' words compared to 11,983 ``female'' words. 
We randomly sampled 30 words from the range spanning the mean of the ``male" distribution to two standard deviations (StD) left and three StD right, excluding male names (e.g., Bob). The same process was mirrored for ``female" words using the inverse bounds. This resulted in a total of 60 words.

Using ChatGPT, we generated sentences with a consistent structure, inserting a male- or female-colored word: ``{\em I am $\langle$colored word$\rangle$. It defines how I approach life and interact with the world}''. For example, ``{\em I am a provider. It defines how I approach life and interact with the world}''. This uniform structure ensured that only the gender-colored word varied, keeping all other elements constant.

\subsection{Baselines}
To provide a basis for comparison, we designed two baselines: (i) ``Professions with Names'' to test whether Bark has any gender knowledge, and (ii) neutral texts for evaluating the distribution of genders in Bark's output for a neutral text. \\

\begin{figure}
    \centering
    \includegraphics[width=1\linewidth]{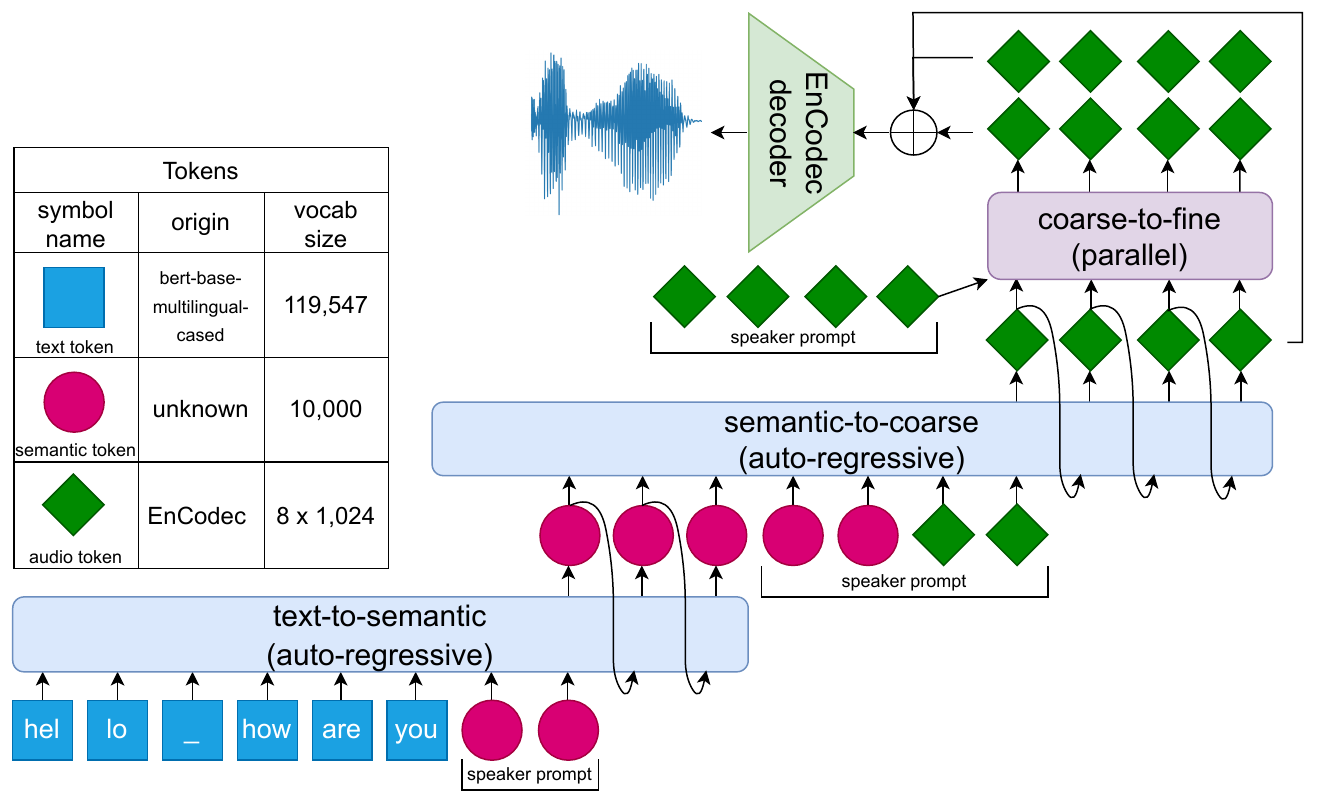}
    \caption{Bark's architecture from \cite{wang2024evaluatingtexttospeechsynthesislarge}}
    \label{fig:bark-architecture}
\end{figure}

\noindent\textbf{Baseline 1: Professions with Names.} To assess Bark's gender awareness, we designed an experiment to test whether it consistently assigns a speaker whose gender aligns with a gender-conforming name. For this purpose, we used sentences from the Professions dataset and incorporated names into them. For example: ``\textit{My name is David. I work as a developer $\cdots$''}. \\

\noindent\textbf{Baseline 2: Neutral Texts.} To analyze the distribution of male and female voices produced by Bark in a neutral context, we used 10 neutral texts: (i) two phonetics-standard passages—the ``Grandfather Passage" and the ``Rainbow Passage"—and (ii) eight Wikipedia abstracts on diverse topics such as education, geography, and politics. Each text was split into sentences to fit Bark’s 12-second audio limit, with each sentence processed 10 times to match the test dataset conditions.

\subsection{Testing with speaker prompt} \label{speaker-prompt-subsection}
Bark consists of three layers, as illustrated in Figure~\ref{fig:bark-architecture}: (i) the text-to-semantic layer is an autoregressive transformer that outputs semantic tokens for the text input and additionally encodes speaker's prompt (if given); (ii) the semantic-to-coarse layer takes in semantic tokens and outputs coarse tokens (which in turn are from EnCodec model~\cite{encodec}); and (iii) the coarse-to-fine layer takes the coarse tokens and outputs fine tokens, which then are turned into a waveform with the EnCodec decoder~\cite{wang2024evaluatingtexttospeechsynthesislarge}. 

Bark provides 10 English speaker prompts--nine male and one female\footnote{https://tinyurl.com/nhffse68}. If no prompt is selected, Bark determines the voice by itself. The sentences from the datasets mentioned above were input without a prompt. The baselines were also tested without a speaker prompt. If a prompt is chosen, it must pass through all model layers, including text-to-semantic. Since Bark claims to capture emotional cues from text, it is worth examining whether it also picks up gender cues and how the text-to-semantic layer influences gender of an assigned speaker.  

To investigate this, we design an experiment where the model receives text input with a speaker prompt that bypasses the text-to-semantic layer but passes through subsequent layers. This modification is tested with sentences from Baseline 1 and Baseline 2 (Rainbow Passage) in (i) a neutral setting (no prompt) and in (ii) a prompt setting (one male and one female prompt, bypassing and passing through text-to-semantic).  

We hypothesize that in a neutral setting, names in the input from Baseline 1 would determine gender of an assigned speaker. Moreover, cases of gender contradiction (e.g., a female-associated name like \textit{Anna} in the sentence but the text-to-sematic layer is prompted with a male speaker prompt) could reveal how much text-to-semantic layer influences the gender of an assigned speaker and whether the model’s inference of gender from text could overcome the speaker prompt. A significant difference in the gender distribution between passing through and bypassing this layer would indicate its impact. We additionally include Baseline 2 since its text input lacks gender inclination, allowing us to observe the distribution of genders for assigned speakers under neutral conditions. Thus testing with Baseline 1 sentences without speaker prompt in the text-to-semantic layer could verify this hypothesis.

\section{Experiments and Results}

\subsection{Baselines}

\noindent\textbf{Baseline 1: Professions with Names.}
This baseline model achieves high precision and recall, as demonstrated in Table~\ref{tab:precision_recall}. We define ground truth based on names and their expected gender (e.g., ``David'' as male and ``Anna'' as female). A sentence with a gender-conforming name should result in an utterance with the corresponding gender of an assigned speaker. The high precision and recall demonstrate that Bark is gender-aware in the names setting. \\

\begin{figure}
    \centering
    \includegraphics[width=0.8\linewidth]{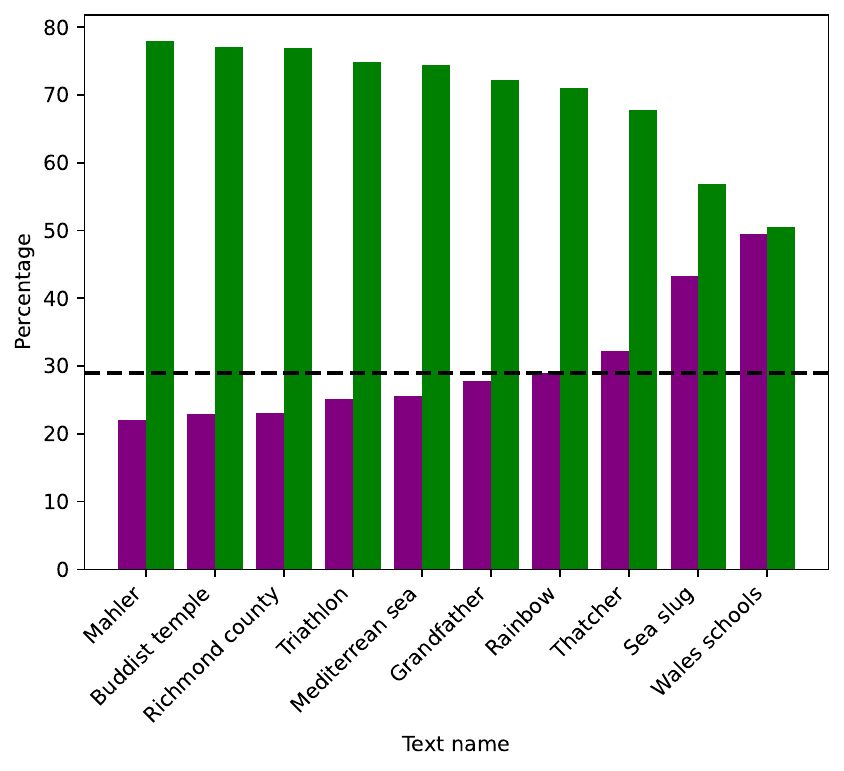}
    \caption{Distribution of female (purple)/male (green) outputs from Bark in a neutral setting. The blue line shows the distribution for the chosen Baseline 2 (Rainbow Passage).}
    \label{fig:baseline2}
\end{figure}

\noindent\textbf{Baseline 2: Neutral Texts}. As shown in Figure~\ref{fig:baseline2}, the majority of the texts have a female voice proportion of up to 30\%. Exceptions include ``Wales schools" and ``Sea slug", based on Wikipedia articles about the Welsh educational system and a sea creature, respectively. For subsequent tests, we assume Rainbow Passage’s female-to-male proportion of 29:71 as the model’s gender distribution without speaker 
prompt on neutral text. 
However, as Table~\ref{tab:ols-regression} demonstrates, when controlling for text and prompt input, the base probability (intercept) of a female utterance increases to 45\%. We define this as the neutrality threshold for each text input in both test datasets.

\begin{table}[t]
\small
\caption{Precision and recall for the Professions and GCW datasets in comparison with Baseline 1.}
\label{tab:precision_recall}
\centering
\setlength{\tabcolsep}{4pt}
\renewcommand{\arraystretch}{1.1}
\begin{tabular}{lcccc}
\toprule
         & \textbf{Baseline 1} & \textbf{Professions} & \textbf{GCW} \\ 
\midrule
\textbf{Prec. (f)} &  0.88 & 0.5 &  0.56 \\
\textbf{Recall (f)} & 0.8   & 0.6 &    0.66  \\
\textbf{Prec. (m)} &  0.81 & 0.49 &  0.58  \\
\textbf{Recall (m)} & 0.89   & 0.38 &   0.5   \\
\bottomrule
\end{tabular}
\end{table}

\subsection{Datasets Results}  

Table~\ref{tab:precision_recall} shows that precision and recall for the Professions dataset are near chance level (50\%) or even lower (38\% for male recall), while the GCW dataset has slightly higher values. However, the precision and recall for both datasets are significantly lower than those of Baseline 1. Precision and recall are calculated for each datapoint where the true value is considered an expected bias (\textit{``nurse"} is female or \textit{``chivalry"} is male for Professions and GCW datasets respectively).

\begin{figure}
    \centering
    \includegraphics[width=0.8\linewidth]{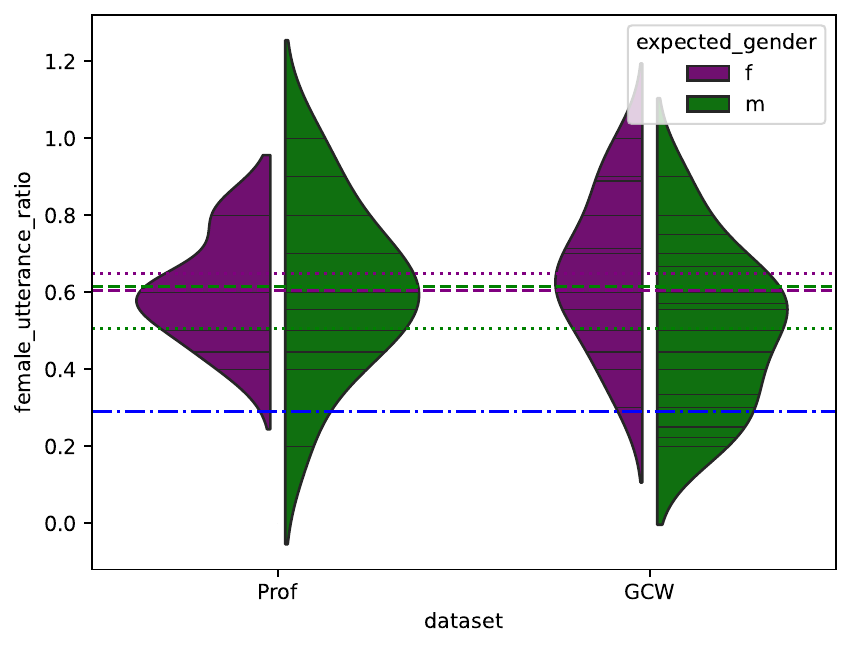}
    \caption{The distribution of female speakers in the Professions (Prof) and GCW datasets. The dashed and dotted lines indicate the mean in the Professions and GCW datasets, respectively, while the blue line represents the mean for Baseline 2.}
    \label{fig:violin-distributions}
\end{figure}

Figure \ref{fig:violin-distributions} demonstrates the distribution of female utterance ratios for female- and male-parts in both datasets. For the Professions dataset, the assigned speaker gender for both groups is centered around 0.6. The overlap between female-expected and male-expected parts suggests that the model's predictions do not strongly distinguish between genders, which is confirmed by a high p-value (1.0) from the Mann-Whitney U Test.

For the GCW dataset, the female and male distributions have means of 0.65 and 0.5, respectively, with a significant difference according to the Mann-Whitney U Test (p-value = 0.007). The female distribution skews higher, and the male distribution skews lower, suggesting a slightly stronger alignment between expected bias and assigned speaker than in the Professions dataset. This aligns with the higher precision and recall values for the GCW dataset, as shown in Table~\ref{tab:precision_recall}. However, significance testing between the Professions and GCW distributions using the Mann-Whitney U Test yields a high p-value (0.32), indicating no meaningful difference. The substantial overlap between female and male distributions in both datasets suggests that Bark does not strongly differentiate gender in text input. \\

\noindent\textbf{Professions Dataset Results} demonstrate diversity for speaker assignment for 65\% of professions (17/26). The remaining results are either female-inclined (female utterance ratio $\ge$ 0.7) or male-inclined (female utterance ratio $\le$ 0.3). Of the nine gender-inclined professions, eight are female-inclined, with six remaining consistent across two runs. These words are shown in Table~\ref{tab:consistent-words}. While some professions preferred by Bark align with stereotypes, others do not. The only profession consistently assigned a male speaker across two runs is \textit{``mechanic”}, reflecting its stereotypically male association.

\noindent\textbf{GCW Dataset Results} demonstrate that 30 out of 60 gender-colored words (50\%) are neutral. The remaining 50\% of words reflect Bark's gender preferences in speaker assignment, being either female-inclined (female utterance ratio $\geq$ 0.7) or male-inclined (female utterance ratio $\leq$ 0.3). Notably, 67\% (20/30) of those are female-inclined, with most (70\%) containing stereotypically female-colored words (\textit{``tutu”}, \textit{``belle”}) but also some male-colored words (\textit{``gallant”}, \textit{``handsome”}). However, only eight words, as shown in Table~\ref{tab:consistent-words}, consistently maintained the same preference across two runs. Similarly, among 10 male-inclined words—including both female- and male-colored ones (\textit{“hips”} and \textit{“mustache”}, respectively)—only two were consistently spoken by a male speaker across both runs.

\begin{table}[t]
\caption{Words with gender preferences (70\%+ utterances), consistent across two runs
}
\label{tab:consistent-words}
\centering
\setlength{\tabcolsep}{0.05\columnwidth} 
\begin{tabular*}{\columnwidth}{l@{\extracolsep{\fill}}cc|l@{\extracolsep{\fill}}cc}
\toprule
\multicolumn{3}{c}{\textit{\textbf{Professions}}} & \multicolumn{3}{c}{\textit{\textbf{GCW}}} \\
\midrule
\textbf{Word} & \textbf{Bias} & \textbf{Pick} & \textbf{Word} & \textbf{Bias} & \textbf{Pick} \\
\midrule
Mechanic & m & m & Guy & m & m \\
Analyst & m & f & Macho & m & m \\
CEO & m & f & Handsome & m & f \\
Cook & m & f & Seashell & f & f \\
Hairdresser & f & f & Uptown & f & f \\
Librarian & f & f & Corset & f & f \\
Teacher & f & f & Feminist & f & f \\
- & - & - & Fussy & f & f \\
- & - & - & Prissy & f & f \\
- & - & - & Housewife & f & f \\
\bottomrule
\end{tabular*}
\end{table}

\subsection{Results from testing with speaker prompt in text-to-semantic layer}
As mentioned in \ref{speaker-prompt-subsection}, we tested synthesizing baselines sentences with and without providing speaker prompt to text-to-semantic layer. For significance testing, we build an Ordinary Least Squares (OLS) regression model predicting the probability assigned to the outcome being a female voice from the text gender, speaker prompt and use of the text2semantic layer as well as their interaction effects on the combined dataset for the two baselines. Neutral or non relevant parameters are coded as 0, female- and male-inclined settings as 1 and -1 respectively.
Table~\ref{tab:ols-regression} shows that the model is statistically significant, and can explain up to 69.5\% of the variation in Bark's female speaker assignment. The intercept value suggests that Bark selects a female speaker in 45.9\% of cases in a neutral setting.  

\texttt{text gender} $\times$ \texttt{text2semantic} has a negative interaction, indicating that the effect of text gender diminishes when \texttt{text2semantic} is present. This supports our hypothesis that when text gender (a name) contradicts a speaker prompt, the absence of the text-to-semantic layer gives more influence to text gender. Similarly,  
\texttt{prompt} $\times$ \texttt{text2semantic} shows a positive interaction, indicating that the effect of the speaker prompt is amplified when passed through the text-to-semantic layer.

 \begin{table}[t]
    \centering
    \caption{Significant factors in OLS Regression Results}    
    \label{tab:ols-regression}
    \begin{tabular}{l c c c}
        \hline
        Variable & Coeff. ($\beta$) & Std. Error & p-value \\
        \hline
        Intercept & 0.4586 & 0.005 & $<$0.001 \\
        text gender & 0.1088 & 0.008 & $<$0.001 \\
        prompt & 0.4501 & 0.006 & $<$0.001 \\
        text2semantic & 0.0134 & 0.006 & 0.026 \\
        text gender $\times$ text2sem. & -0.0542 & 0.009 & $<$0.001 \\
        prompt $\times$ text2sem. & 0.0459 & 0.006 & $<$0.001 \\
        \hline
        $R^2$ & \multicolumn{3}{ c}{0.695} \\
        F-statistic & \multicolumn{3}{ c}{848.57} \\
        p-value (F-stat) & \multicolumn{3}{ c}{$<$0.001} \\
        \hline
    \end{tabular}
\end{table}

\section{Discussion}
\subsection{Gender bias in Bark}  
The results for both the Professions and GCW datasets indicate that Bark does not exhibit a strong gender bias, where bias is defined as the \textit{systematic} alignment of gender stereotypes with the gender of an assigned speaker. For the Professions dataset, Bark demonstrates diversity for 73\% of professions, while confirming stereotypes in 15\% of cases --specifically, for one male-stereotyped profession \textit{``mechanic''} and three female-stereotyped professions \textit{``hairdresser''}, \textit{``librarian''}, and \textit{``teacher''}. Interestingly, in 12\% of cases, Bark’s gender assignment is counter-stereotypical with female speaker being assigned to three male-associated professions: \textit{``CEO"}, \textit{``cook"}, and \textit{``analyst"}.  

For GCW words, Bark similarly exhibited diversity, assigning a speaker whose gender aligns with stereotypes in only 17\% of cases. However, distinguishing between gender bias and gender information remains nontrivial, as words like \textit{``housewife''} and \textit{``guy''} inherently carry gendered meanings\footnote{ https://dictionary.cambridge.org/uk/dictionary/english/housewife, \\ https://dictionary.cambridge.org/uk/dictionary/english/guy}, however, the systematic association of a female speaker with words like (\textit{``seashell''}, \textit{``uptown''}, \textit{``corset''}, \textit{``prissy''}, \textit{``fussy''}) do show a bias, especially because the latter two words carry out a negative connotation. We believe that a Speech-LLM should not exhibit this bias and reinforce gender associations. 
Our conclusion that Bark does not exhibit strong gender bias is based on its diverse speaker assignments for most words in both datasets. While certain preferences emerged in both the Professions and GCW datasets, they aligned with both stereotypes and counter-stereotypes. However, given the risk of harmful biases in LLMs, it remains crucial to stay vigilant.

\subsection{Limitations}
Both Bark and the SGR model operate on a binary gender framework, either generating or differentiating between only male and female categories. This inherently simplifies the complexity of gender identity and excludes non-binary representations.
This study focuses exclusively on English, despite Bark’s multilingual capabilities. Investigating whether Bark exhibits similar gender biases in other languages would be valuable. For instance, Slavic languages (such as Polish and Russian, which Bark supports) encode gender more explicitly through adjective- and verb-alignment.
The words for GCW dataset, derived from \cite{du-etal-2019-exploring}, are thematically diverse, including terms related to body parts (\textit{``hips''}), identity (\textit{``frat''}), and clothing (\textit{``corset''}). This variability may introduce inconsistencies in bias evaluation.  

\section{Conclusions}
In this paper, we evaluated a Speech-LLM-based TTS model Bark, for the presence of gender bias. We created two text datasets as TTS input to probe potential bias: {\em Professions} and {\em Gender-Colored Words (GCW)}. While Bark did not exhibit systematic gender bias, it displayed a few gender preferences in both settings, including anti- and pro-bias ones, and demonstrated gender awareness in the context of names. We also found proof that Bark does infer gender from the text, specifically, while encoding text input into semantic tokens.

Since speech inherently requires an authored voice, speaker selection in Speech-LLMs provides a direct way to observe gender associations in generative models. While Bark does not exhibit strong systematic bias under the tested conditions, the autonomous assignment of speaker gender by Speech-LLMs could serve as a diagnostic tool, especially if leveraged in model development to uncover underlying patterns in training data that might otherwise go unnoticed.

\section{Acknowledgments}
This research is supported by the Swedish Research Council project Perception of speaker stance (VR-2020-02396), the Riksbankens Jubileumsfond project CAPTivating (P20-0298).
\bibliographystyle{IEEEtran}
\bibliography{mybib}

\begin{thebibliography}{10}
\providecommand{\url}[1]{#1}
\csname url@samestyle\endcsname
\providecommand{\newblock}{\relax}
\providecommand{\bibinfo}[2]{#2}
\providecommand{\BIBentrySTDinterwordspacing}{\spaceskip=0pt\relax}
\providecommand{\BIBentryALTinterwordstretchfactor}{4}
\providecommand{\BIBentryALTinterwordspacing}{\spaceskip=\fontdimen2\font plus
\BIBentryALTinterwordstretchfactor\fontdimen3\font minus \fontdimen4\font\relax}
\providecommand{\BIBforeignlanguage}[2]{{%
\expandafter\ifx\csname l@#1\endcsname\relax
\typeout{** WARNING: IEEEtran.bst: No hyphenation pattern has been}%
\typeout{** loaded for the language `#1'. Using the pattern for}%
\typeout{** the default language instead.}%
\else
\language=\csname l@#1\endcsname
\fi
#2}}
\providecommand{\BIBdecl}{\relax}
\BIBdecl

\bibitem{zhao2024genderbiaslargelanguage}
J.~Zhao, Y.~Ding, C.~Jia, Y.~Wang, and Z.~Qian, ``Gender bias in large language models across multiple languages,'' \emph{arXiv e-prints}, pp. arXiv--2403, 2024.

\bibitem{Kotek_2023}
\BIBentryALTinterwordspacing
H.~Kotek, R.~Dockum, and D.~Sun, ``Gender bias and stereotypes in large language models,'' in \emph{Proceedings of The ACM Collective Intelligence Conference}, ser. CI ’23.\hskip 1em plus 0.5em minus 0.4em\relax ACM, Nov. 2023. [Online]. Available: \url{http://dx.doi.org/10.1145/3582269.3615599}
\BIBentrySTDinterwordspacing

\bibitem{knowledge-distillation}
J.~Ahn, H.~Lee, J.~Kim, and A.~Oh, ``Why knowledge distillation amplifies gender bias and how to mitigate from the perspective of distilbert,'' in \emph{Proceedings of the 4th Workshop on Gender Bias in Natural Language Processing (GeBNLP)}, 2022, pp. 266--272.

\bibitem{lakhotia-etal-2021-generative}
\BIBentryALTinterwordspacing
K.~Lakhotia, E.~Kharitonov, W.-N. Hsu, Y.~Adi, A.~Polyak, B.~Bolte, T.-A. Nguyen, J.~Copet, A.~Baevski, A.~Mohamed, and E.~Dupoux, ``On generative spoken language modeling from raw audio,'' \emph{Transactions of the Association for Computational Linguistics}, vol.~9, pp. 1336--1354, 2021. [Online]. Available: \url{https://aclanthology.org/2021.tacl-1.79}
\BIBentrySTDinterwordspacing

\bibitem{betker2023better}
J.~Betker, ``Better speech synthesis through scaling,'' \emph{arXiv e-prints}, pp. arXiv--2305, 2023.

\bibitem{lajszczak2024base}
M.~Lajszczak, G.~C. Ruiz, Y.~Li, F.~Beyhan, A.~van Korlaar, F.~Yang, A.~Joly, {\'A}.~M. Cortinas, A.~Abbas, A.~Michalski \emph{et~al.}, ``Base tts: Lessons from building a billion-parameter text-to-speech model on 100k hours of data,'' 2024.

\bibitem{tathe2024end}
A.~Tathe, A.~Kamble, S.~Kumbharkar, A.~Bhandare, and A.~C. Mitra, ``End to end hindi to english speech conversion using bark, mbart and a finetuned xlsr wav2vec2,'' \emph{arXiv e-prints}, pp. arXiv--2401, 2024.

\bibitem{kamble2024customdataaugmentationlow}
\BIBentryALTinterwordspacing
A.~Kamble, A.~Tathe, S.~Kumbharkar, A.~Bhandare, and A.~C. Mitra, ``Custom data augmentation for low resource asr using bark and retrieval-based voice conversion,'' 2024. [Online]. Available: \url{https://arxiv.org/abs/2311.14836}
\BIBentrySTDinterwordspacing

\bibitem{schumacher2023enhancing}
D.~Schumacher, F.~LaBounty~Jr, and M.~Schellekens, ``Enhancing suno's bark text-to-speech model: Addressing limitations through meta's encodec and pre-trained hubert,'' \emph{Available at SSRN 4443815}, 2023.

\bibitem{kim2021conditional}
J.~Kim, J.~Kong, and J.~Son, ``Conditional variational autoencoder with adversarial learning for end-to-end text-to-speech,'' in \emph{International Conference on Machine Learning}.\hskip 1em plus 0.5em minus 0.4em\relax PMLR, 2021, pp. 5530--5540.

\bibitem{wang2024evaluatingtexttospeechsynthesislarge}
S.~Wang and {\'E}.~Sz{\'e}kely, ``Evaluating text-to-speech synthesis from a large discrete token-based speech language model,'' in \emph{Proceedings of the 2024 Joint International Conference on Computational Linguistics, Language Resources and Evaluation (LREC-COLING 2024)}, 2024, pp. 6464--6474.

\bibitem{madspeech}
\BIBentryALTinterwordspacing
M.~Futeral, A.~Agostinelli, M.~Tagliasacchi, N.~Zeghidour, and E.~Kharitonov, ``Mad speech: Measures of acoustic diversity of speech,'' 2024. [Online]. Available: \url{https://arxiv.org/abs/2404.10419}
\BIBentrySTDinterwordspacing

\bibitem{Székely2023ProsodycontrollableGS}
\BIBentryALTinterwordspacing
{\'E}.~Sz{\'e}kely, J.~Gustafson, and I.~Torre, ``Prosody-controllable gender-ambiguous speech synthesis: A tool for investigating implicit bias in speech perception,'' in \emph{Interspeech}, 2023. [Online]. Available: \url{https://api.semanticscholar.org/CorpusID:260906437}
\BIBentrySTDinterwordspacing

\bibitem{baevski2020wav2vec}
A.~Baevski, Y.~Zhou, A.~Mohamed, and M.~Auli, ``wav2vec 2.0: A framework for self-supervised learning of speech representations,'' in \emph{Advances in Neural Information Processing Systems}, vol.~33, 2020, pp. 12\,449--12\,460.

\bibitem{libritts}
H.~Zen, V.~Dang, R.~Clark, Y.~Zhang, R.~J. Weiss, Y.~Jia, Z.~Chen, and Y.~Wu, ``Libritts: A corpus derived from librispeech for text-to-speech,'' \emph{Interspeech 2019}, 2019.

\bibitem{winobias}
J.~Zhao, T.~Wang, M.~Yatskar, V.~Ordonez, and K.-W. Chang, ``Gender bias in coreference resolution: Evaluation and debiasing methods,'' in \emph{Proceedings of the 2018 Conference of the North American Chapter of the Association for Computational Linguistics: Human Language Technologies}, vol.~2, 2018.

\bibitem{du-etal-2019-exploring}
\BIBentryALTinterwordspacing
Y.~Du, Y.~Wu, and M.~Lan, ``Exploring human gender stereotypes with word association test,'' in \emph{Proceedings of the 2019 Conference on Empirical Methods in Natural Language Processing and the 9th International Joint Conference on Natural Language Processing (EMNLP-IJCNLP)}, K.~Inui, J.~Jiang, V.~Ng, and X.~Wan, Eds.\hskip 1em plus 0.5em minus 0.4em\relax Hong Kong, China: Association for Computational Linguistics, Nov. 2019, pp. 6133--6143. [Online]. Available: \url{https://aclanthology.org/D19-1635}
\BIBentrySTDinterwordspacing

\bibitem{encodec}
A.~D{\'e}fossez, J.~Copet, G.~Synnaeve, and Y.~Adi, ``High fidelity neural audio compression,'' \emph{Transactions on Machine Learning Research}, 2023.

\end{thebibliography}

\end{document}